\documentclass[acmtog,nonacm]{acmart}

\AtBeginDocument{%
  }
\usepackage{amsmath}
\usepackage{multirow}
\usepackage{placeins}

\setcopyright{none}
\acmDOI{}
\acmISBN{}
\begin{document}

\title{PhysReflect: Geometry and Perception Guided Diffusion for Physically-Plausible Mirror Reflections}

\author{Shuheng Ge}
\email{suheng@hit.edu.cn}
\affiliation{%
  \institution{Harbin Institute of Technology}
  \city{Harbin}
  \country{China}
}

\author{Hongwei Ren}
\affiliation{%
	\institution{Harbin Institute of Technology}
	\city{Harbin}
	\country{China}}
\email{Renhongwei@hit.edu.cn}

\author{Li Zhang}
\affiliation{%
	\institution{Harbin Institute of Technology}
	\city{Harbin}
	\country{China}}
\email{zhangli92@hit.edu.cn}

\author{Xiangqian Wu}
\affiliation{%
  \institution{Harbin Institute of Technology}
\city{Harbin}
\country{China}}
\email{xqwu@hit.edu.cn}

\renewcommand{\shortauthors}{Ge et al.}
\begin{abstract}
Diffusion models generate high-quality images, yet often violate the physical laws governing mirror reflections.
Reflections often suffer from geometric aberrations, including positional offsets, directional misalignment, proportional imbalance, and structural distortion. These failures remain evident even in contemporary state-of-the-art generative systems. Existing methods mitigate this problem through synthetic data scaling or auxiliary depth conditioning, yet their reliance on latent-space denoising objectives leaves reflection-specific geometric and perceptual constraints only indirectly enforced.
To bridge this gap, we present \textbf{PhysReflect}, a geometry- and perception-guided diffusion framework that decodes the predicted clean latent into image space at each training step, where two complementary mirror-specific objectives are applied under a reliability-aware annealed supervision scheme.

The \textbf{Geometric Loss} enforces mirror-induced spatial consistency through sparse epipolar correspondence and dense boundary projection alignment, where a SAM2-based \textit{TwinTrack} mechanism provides stable in-mirror localization for boundary-aware supervision. 
The \textbf{Perceptual Loss} preserves reflected appearance by combining \textbf{Semantic Consistency Loss}, which maintains reflected identity and appearance via DINOv2 features, and \textbf{Lighting Consistency Loss}, which regularizes depth, surface-normal, and illumination coherence under monocular geometry priors. 
Experiments on synthetic and real-world benchmarks show that PhysReflect outperforms prior mirror-reflection methods in geometric, perceptual, and physical-plausibility metrics, as well as qualitative visual results.

\end{abstract}



\begin{teaserfigure}
\centering
	\includegraphics[width=\textwidth]{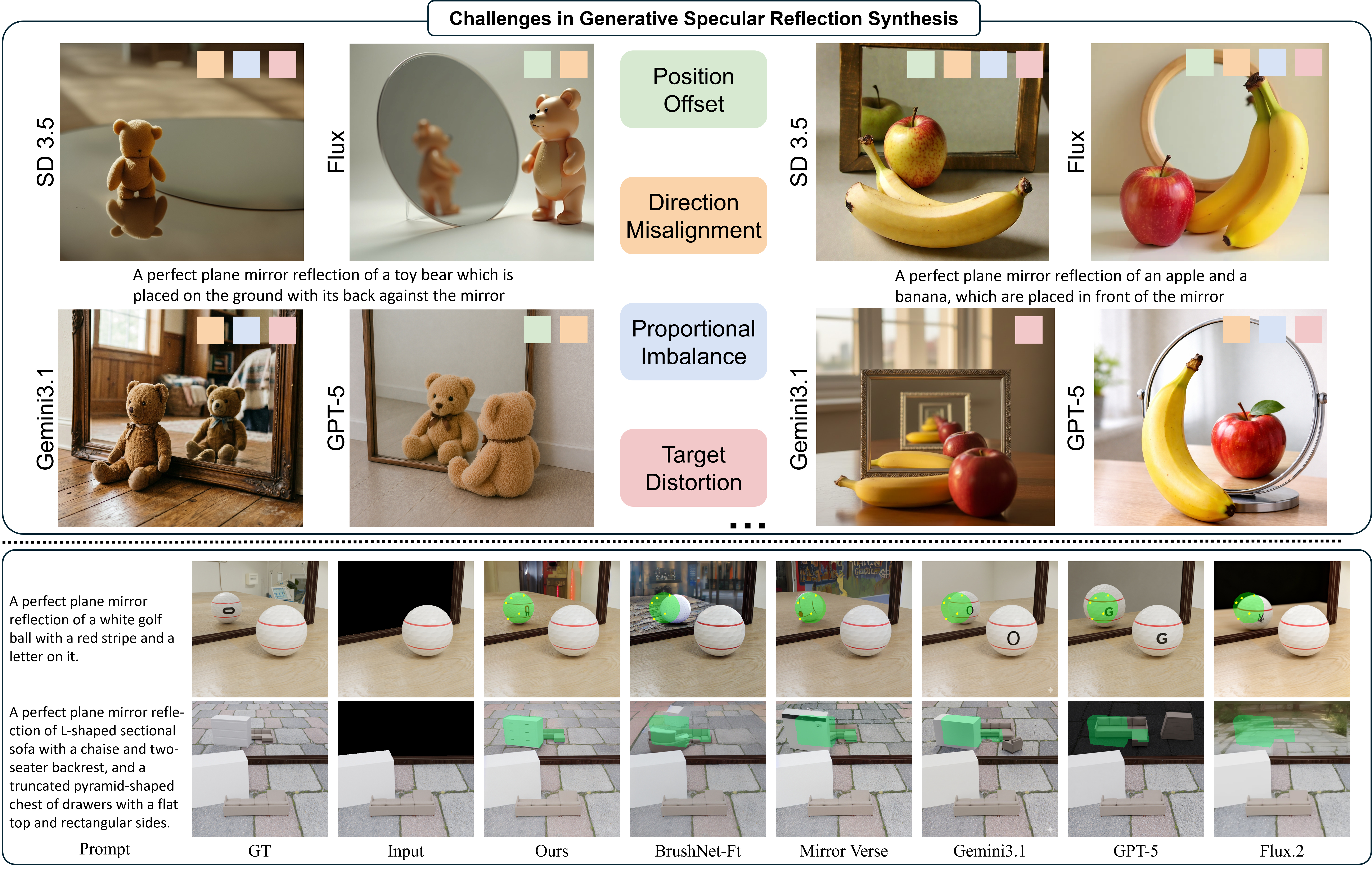}
	\caption{(Top) \textbf{Challenges in Generative Specular Reflection Synthesis.} State-of-the-art open-source models (Stable Diffusion 3.5~\cite{sd35}, Flux~\cite{flux}) and commercial models (GPT-5, Gemini Pro 3.1) consistently produce physically implausible mirror reflections with geometric and perceptual inconsistencies. (Bottom) \textbf{Comparison with existing methods and commercial models.} The green mask overlaid on each example indicates the geometrically correct in-mirror object location dictated by the reflection geometry, providing a reference against which the positional and shape deviations of the generated reflections become directly apparent. Our PhysReflect generates reflections that are superior in geometric orientation, spatial positioning, and semantic alignment with the source objects.}
	\label{fig:teaser}
\end{teaserfigure}

\maketitle



\section{Introduction}
\label{sec:intro}
Diffusion-based generative models have become a standard foundation for visual synthesis and editing, supported by strong progress in image generation~\cite{ho2020ddpm, rombach2022stablediffusion, podell2023sdxl}, controllable generation~\cite{controlnet, mo2024freecontrol, parihar2024precisecontrol, parihar2025text2place}, and image inpainting~\cite{brushnet}. 
Despite their strong visual realism, prior studies have reported systematic failures in shadows~\cite{sarkar2024shadows}, perspective geometry~\cite{upadhyay2023perspective}, and specular effects~\cite{objectdrop}. 
Mirror reflection presents a particularly challenging case because a plausible reflected scene must satisfy 
coupled geometric and perceptual constraints.
The reflected content must appear at geometrically correct locations while preserving the identity and appearance of the source object and maintaining coherent illumination with the surrounding scene.
As shown in Figure~\ref{fig:teaser}, although recent open-source models such as Stable Diffusion 3.5~\cite{sd35} and Flux~\cite{flux}, as well as commercial systems such as GPT-5 and Gemini Pro, can synthesize high-fidelity visual content, their reflections still exhibit positional offsets, directional misalignment, proportional imbalance, and structural distortion.

To mitigate these reflection inconsistencies, recent methods formulate mirror reflection synthesis as a conditional image inpainting task, relying on large-scale synthetic data, auxiliary geometric conditions, and curriculum-based training strategies to improve reflection realism. MirrorFusion~\cite{mirrorfusion} constructs a large-scale synthetic dataset with depth, normal, and segmentation annotations, and introduces outside-mirror depth maps as conditional inputs to provide stronger geometric cues, encouraging the model to learn reflection regularities.
Building upon this direction, MirrorVerse~\cite{mirrorverse} further builds a larger and higher-quality synthetic dataset, and proposes a three-stage curriculum from single-object scenes to multi-object scenes and then to real-world scenes, improving the ability of the model to handle complex reflection scenarios. 

Despite these advances, these methods still share a common limitation: they lack explicit physical supervision and rely primarily on latent-space denoising objectives, where reflection geometry and perceptual consistency are only indirectly learned from paired data.

We therefore propose \textbf{PhysReflect}, a geometry- and perception-guided diffusion training framework for mirror reflection generation. During training, PhysReflect decodes the predicted clean latent into image space and applies two complementary mirror-specific objectives for geometric and perceptual consistency under reliability-aware annealed supervision.

PhysReflect organizes this supervision into two complementary objectives: 
\textbf{Geometric Loss} ($\mathcal{L}_{geo}$) and 
\textbf{Perceptual Loss} ($\mathcal{L}_{perc}$). 
The Geometric Loss constrains mirror-induced spatial consistency through sparse epipolar correspondence and dense boundary projection alignment. 
Within this geometric objective, TwinTrack, a SAM2-based dual-frame propagation mechanism, provides stable in-mirror localization for boundary-aware supervision under generated positional drift and shape deformation.
The Perceptual Loss preserves reflected appearance by combining semantic consistency for source identity and appearance preservation with lighting consistency for depth, surface-normal, and illumination coherence.

We validate PhysReflect on both synthetic MirrorBenchV2~\cite{mirrorverse} and real-world MSD~\cite{msd} mirror datasets against the state-of-the-art mirror reflection baseline MirrorVerse. To better evaluate whether generated reflections satisfy geometric and perceptual reflection constraints, we extend the standard image-quality metrics with additional evaluation protocols, including DINO Similarity for perceptual consistency, Object IoU and Boundary F1 for geometric alignment, and SH Err for illumination coherence. Quantitative and qualitative results show that, compared with existing methods, PhysReflect substantially improves the physical plausibility of generated mirror reflections while preserving image quality. Since all geometric and perceptual losses and auxiliary models are used only during training, PhysReflect does not introduce additional inference-time cost.

Our contributions are summarized as follows:
\begin{itemize}
    \item We present \textbf{PhysReflect}, a geometry- and perception-guided training framework for physically plausible mirror reflection generation. Building on established decoded-output supervision, PhysReflect formulates and integrates  mirror-specific geometric and perceptual objectives while leaving the baseline inference pipeline unchanged.
    
    \item We introduce a \textbf{Geometric Loss} that enforces mirror-induced spatial consistency through sparse epipolar correspondence and dense boundary projection alignment. A SAM2-based \textit{TwinTrack} mechanism is integrated into this geometric objective to provide stable in-mirror localization for boundary-aware supervision.

    \item We adapt pretrained-feature perceptual supervision to mirror reflection using global and mirror-local DINOv2 features to regularize source-to-reflection appearance and semantic consistency, together with geometry-aware low-frequency illumination regularization for illumination coherence.
    
    \item Experiments on synthetic and real-world mirror datasets show that PhysReflect improves geometric accuracy, perceptual fidelity, illumination coherence, and physical plausibility over existing mirror-reflection generation methods.
\end{itemize}
\section{Related Work}

\subsection{Diffusion Inpainting and Conditional Control}

Diffusion models and their latent variants have become the dominant backbone for image synthesis and editing~\cite{ho2020ddpm,rombach2022stablediffusion}. For inpainting, prior work either modifies the sampling process of pre-trained models~\cite{avrahami2023blended,lugmayr2022repaint} or trains task-specific architectures that condition on masks, masked images, text, or spatial cues~\cite{brushnet,controlnet}. 
These approaches are effective at improving visual fidelity and controllability, but their constraints are usually injected as inputs or architectural conditions. Such conditioning can guide where and what to generate, yet it does not enforce whether the completed content satisfies task-specific physical relations.

This distinction is central to mirror reflection generation: conditioning provides structural hints, whereas 
reflection-specific geometric and perceptual consistency must be assessed on the generated pixels.
PhysReflect therefore complements conditional control with output-space losses for the decoded prediction.

\subsection{Mirror Reflection Generation}

Mirror reflection is a long-standing challenge in visual perception and scene understanding. Prior work has mainly studied it from the perspectives of reflective-surface understanding, including mirror or glass region detection and segmentation~\cite{msd,lin2020pmd,mei2022glass}, and 3D scene reconstruction, where mirror-like effects are modeled with specialized NeRF or 3D Gaussian representations~\cite{msnerf,mirror3dgs,mirrorgaussian}. These reconstruction-oriented methods are designed for scene-level 3D modeling and usually require multi-view observations, camera poses, and iterative optimization. Recently, with the development of generative models, mirror reflection synthesis has emerged as a challenging setting for studying whether visual generation models can follow physical regularities from lightweight image-level inputs.

MirrorFusion~\cite{mirrorfusion} formulates mirror reflection synthesis as a conditional inpainting problem, introducing SynMirror and a depth-conditioned BrushNet baseline for paired-data reflection generation. MirrorVerse~\cite{mirrorverse} extends this direction with SynMirrorV2, diverse layouts, multi-object scenes, and a curriculum for improved generalization to multi-object and real-world scenes. However, these methods still rely primarily on data-driven reconstruction objectives. Depth conditioning and inpainting masks provide structural cues, but under the standard latent denoising objective, geometric constraints such as source-reflection correspondence and boundary alignment, as well as perceptual constraints such as semantic and illumination consistency 
remain only indirectly constrained.
PhysReflect targets this gap by converting these reflection-specific consistency requirements into differentiable training losses, while keeping the inference-time inputs unchanged.

\subsection{Differentiable Supervision and Geometric Priors}

Recent studies show that visually realistic generative models can still violate physical and geometric regularities such as shadows, perspective, and specular effects~\cite{sarkar2024shadows,upadhyay2023perspective,objectdrop}. Since these properties are directly reflected in the generated image, prior work has explored differentiable image- and feature-level supervision beyond the standard latent denoising objective. ReFL~\cite{xu2023imagereward} evaluates a one-step clean-image estimate with a learned reward, while AlignProp~\cite{prabhudesai2023alignprop} and DRaFT~\cite{clark2024draft} backpropagate differentiable rewards through the denoising or sampling process. REPA~\cite{yu2025repa} aligns denoiser representations with pretrained visual features, and PixelGen~\cite{ma2026pixelgen} applies perceptual and DINO-based losses to clean-image predictions. These works establish decoded-output and pretrained-feature supervision as general techniques; PhysReflect specializes them to mirror reflection through source--reflection geometric constraints and mirror-local appearance and illumination priors.

Physical and geometric priors have also been explored in task-specific generation. Relighting methods incorporate estimated geometry and illumination to improve physical consistency~\cite{pilight,unilumos}, while novel-view synthesis methods exploit camera geometry. Zero-1-to-3~\cite{liu2023zero123} conditions generation on a source image and relative camera pose, whereas EpiDiff~\cite{huang2024epidiff} introduces epipolar-constrained interaction for cross-view feature exchange. Mirror reflection generation, in contrast, requires source-to-reflection correspondence, reflected-boundary alignment, appearance preservation, and illumination coherence within a single image. PhysReflect organizes these requirements into geometric guidance for correspondence and boundary alignment, together with perceptual guidance for semantic and illumination consistency.

\section{Preliminaries and Problem Formulation}
\label{sec:prelim}
Following MirrorFusion~\cite{mirrorfusion}, we formulate mirror reflection synthesis as conditional image inpainting, where the mirror mask defines the generation region and the model fills it with plausible reflection content. We adopt a depth-conditioned BrushNet~\cite{brushnet} backbone, whose Generation Net is a latent diffusion model~\cite{rombach2022stablediffusion} and whose Conditioning Net injects multi-scale features from the masked image, mirror mask, and outside-mirror depth map through zero-convolution layers. During training, the VAE encoder maps the ground-truth image $\mathbf{X}_0$ to $\mathbf{z}_0=\mathcal{E}(\mathbf{X}_0)$, and the model learns the standard noise-prediction objective $\mathcal{L}_{MSE}$ defined in Appendix. The task input is therefore the noised latent $\mathbf{z}_t$, masked image $\mathbf{X}$, mirror mask $\mathbf{M}_{mirror}$, outside-mirror depth $\mathbf{D}$, and text prompt; 
the model predicts the completed image $\hat{\mathbf{X}}_0$, which is supervised by the ground-truth image $\mathbf{X}_0$.

\section{Method}
\begin{figure*}[t]
\centering
\includegraphics[width=\textwidth]{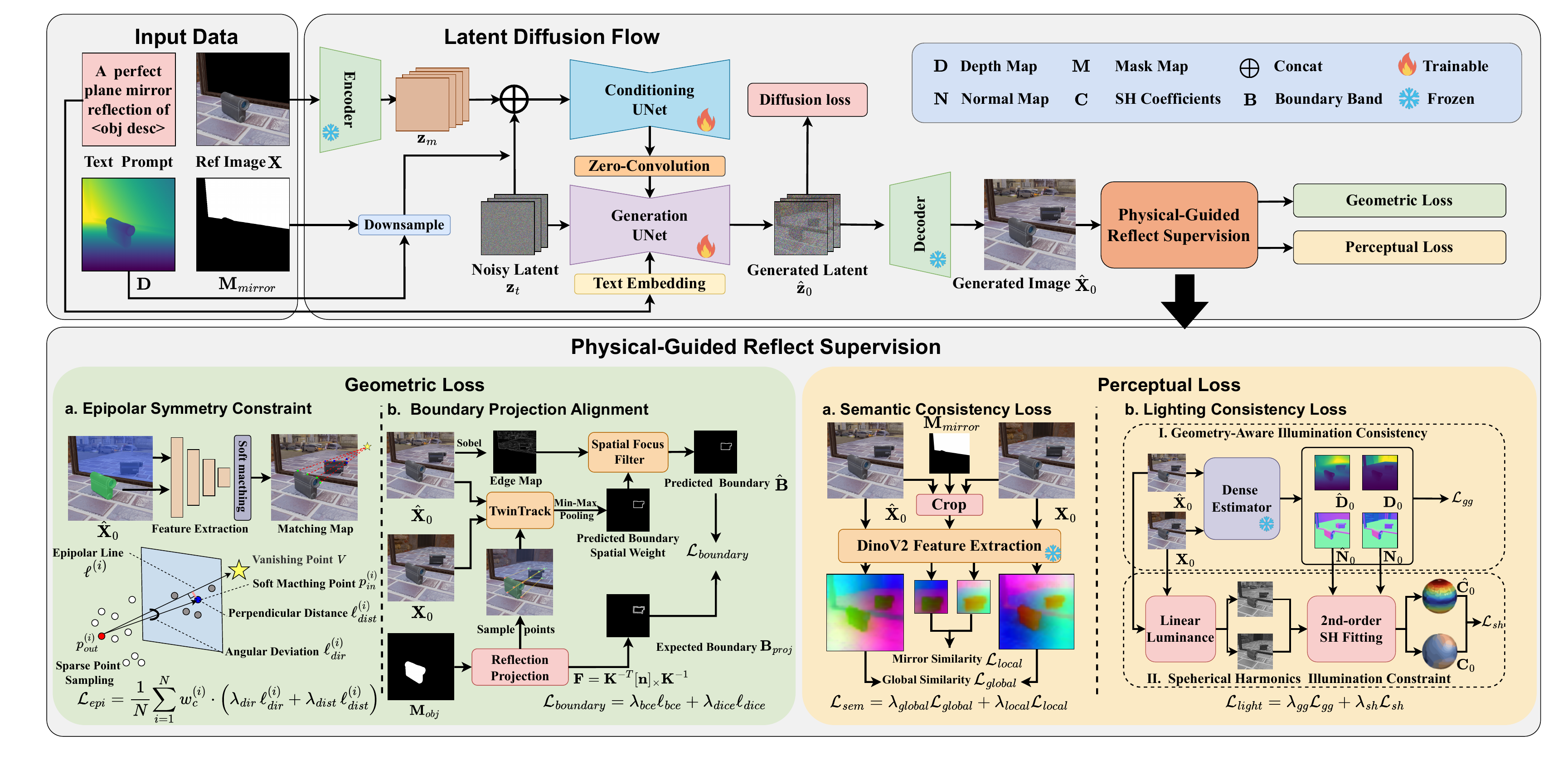}
\caption{\textbf{Overview of PhysReflect.} PhysReflect builds on a depth-conditioned diffusion inpainting backbone and adds mirror-specific physical supervision applied to the decoded prediction in pixel space during training. Given a noised latent, the model predicts the clean latent, decodes it into image space, and computes 
two complementary objectives: Geometric Loss and Perceptual Loss. Geometric Loss enforces mirror-induced spatial consistency, while Perceptual Loss preserves reflected appearance through semantic and lighting consistency.
Auxiliary priors and frozen models construct supervision only during training, while inference uses the original Baseline inputs and sampling graph.
}
\label{fig:framework}
\end{figure*}

\subsection{PhysReflect Overview}
\label{sec:overview}

As shown in Fig.~\ref{fig:framework}, PhysReflect introduces explicit mirror-specific physical supervision, applied to the decoded prediction in pixel space during training. Unlike conventional latent-space denoising objectives, which only indirectly encourage physical consistency through paired training data, PhysReflect evaluates whether generated reflections satisfy \textbf{geometric and perceptual constraints} directly in image space. 

To evaluate the proposed mirror-specific supervision on generated content, we decode the predicted clean latent into image space during training, where reflection geometry, object correspondence, boundary alignment, appearance consistency, and lighting coherence become directly measurable.

Given the predicted clean latent representation $\hat{\mathbf{z}}_0$, we obtain the decoded image prediction as:
\begin{equation}
    \hat{\mathbf{X}}_0 = \mathcal{D}(\hat{\mathbf{z}}_0),
\end{equation}
where $\mathcal{D}$ denotes the pretrained VAE decoder. PhysReflect then applies mirror-specific physical supervision to $\hat{\mathbf{X}}_0$ in pixel space through two complementary objectives: \textbf{Geometric Loss} and \textbf{Perceptual Loss}. 
The Geometric Loss combines sparse epipolar correspondence and dense boundary projection alignment, while the Perceptual Loss combines semantic consistency and lighting consistency.
\subsection{Geometric Loss}
\label{sec:geo}

Mirror reflection generation must place reflected content at physically valid in-mirror locations while preserving geometric symmetry. Existing data-driven objectives provide only implicit spatial supervision, leaving common artifacts such as positional drift and orientation errors. 
We therefore introduce the Geometric Loss $\mathcal{L}_{geo}$, which combines sparse epipolar correspondence supervision with dense boundary projection alignment.

The mirror plane is defined in the camera coordinate system by a unit normal $\mathbf{n}$ and signed distance $d$, with $\mathbf{n}^{T}\mathbf{p}+d=0$ for any 3D point $\mathbf{p}$ on the plane. The virtual point of a real point $\mathbf{p}$ is:
\begin{equation}
    \mathbf{p}_{virt}
    =
    \mathbf{p}
    -
    2(\mathbf{n}^{T}\mathbf{p}+d)\mathbf{n}.
    \label{eq:reflect}
\end{equation}
This displacement is parallel to $\mathbf{n}$. For homogeneous image coordinates $\tilde{\mathbf{p}}_{out}$ and $\tilde{\mathbf{p}}_{in}$ of a real point and its reflected virtual point, the two camera rays and $\mathbf{n}$ lie in one epipolar plane, yielding:
\begin{equation}
    \tilde{\mathbf{p}}_{in}^{T}
    \mathbf{F}
    \tilde{\mathbf{p}}_{out}=0,
    \qquad
    \mathbf{F}=\mathbf{K}^{-T}[\mathbf{n}]_{\times}\mathbf{K}^{-1}.
    \label{eq:fundamental}
\end{equation}
Here $[\mathbf{n}]_{\times}$ is the skew-symmetric matrix of $\mathbf{n}$ and $\mathbf{K}$ is the camera intrinsic matrix. The distance $d$ is absent from Eq.~\ref{eq:fundamental} because epipolar geometry depends only on the mirror-normal direction, but it is still used for metric projection in Eq.~\ref{eq:reflect} when constructing target boundaries. The derivation is given in Appendix.

The above geometry provides two complementary ways to supervise generated reflections.  
First, the epipolar relation in Eq.~\eqref{eq:fundamental} supervises sparse source-to-reflection correspondences. 
Second, the metric reflection projection in Eq.~\eqref{eq:reflect} provides dense boundary-level alignment by projecting object contours across the mirror plane. 
We first describe the sparse epipolar constraint, followed by the boundary projection alignment term.
\subsubsection{Epipolar Symmetry Constraint}
\label{sec:epi}

Epipolar Symmetry enforces sparse real-to-reflected correspondences under the mirror-induced geometry. As shown in Fig.~\ref{fig:framework}, we sample $N$ points $\{\mathbf{p}_{out}^{(i)}\}_{i=1}^N$ outside the mirror, find reflected correspondences $\mathbf{p}_{in}^{(i)}$ by DINOv2-based soft matching, and penalize their epipolar distance and reflection-direction inconsistency. The reflection-direction term uses the mirror-normal vanishing point $\mathbf{V}$ as its geometric reference.
DINOv2~\cite{dinov2} patch features are used because mirror reflection weakens hand-crafted descriptors such as SIFT~\cite{lowe2004sift} and ORB~\cite{rublee2011orb}, 
while DINOv2 provides spatially localized semantic features that are more suitable for dense correspondence than global CLIP embeddings.

To keep retained correspondences differentiable, we replace hard argmax matching with a soft expected position over candidate points $\{\mathbf{c}_j\}$ sampled from an epipolar-line band inside the mirror. For each $\mathbf{p}_{out}^{(i)}$, we extract $L_2$-normalized DINOv2 ViT-S/14 features $\mathbf{f}_{out}^{(i)}$ from the real input image and $\mathbf{f}_{in,j}$ from $\hat{\mathbf{X}}_0$ at candidate position $\mathbf{c}_j$. With temperature $\tau = 0.05$, the matched position is:
\begin{equation}
    \mathbf{p}_{in}^{(i)} = \sum_j p_{ij}\, \mathbf{c}_j, \quad p_{ij} = \frac{\exp(\mathbf{f}_{out}^{(i)} \cdot \mathbf{f}_{in,j} / \tau)}{\sum_k \exp(\mathbf{f}_{out}^{(i)} \cdot \mathbf{f}_{in,k} / \tau)}
    \label{eq:softmatch}
\end{equation}
Gradients pass through this soft expectation and the frozen DINOv2 forward pass to $\hat{\mathbf{X}}_0$. Detached confidence and reciprocal filters retain only reliable matches; samples with too few retained matches skip this term. For each retained match, the epipolar loss combines a normalized point-to-line distance $\ell_{dist}^{(i)}$ and a cosine direction penalty $\ell_{dir}^{(i)}$:
\begin{equation}
    \mathcal{L}_{epi}
    =
    \frac{
    \sum_{i=1}^{N}w_c^{(i)}
    \left(
    \lambda_{dist}\ell_{dist}^{(i)}
    +
    \lambda_{dir}\ell_{dir}^{(i)}
    \right)}
    {\sum_{i=1}^{N}w_c^{(i)}+\epsilon}.
    \label{eq:epi_loss}
\end{equation}

\subsubsection{Boundary Projection Alignment with TwinTrack}
\label{sec:boundary}

The epipolar term constrains sparse point correspondences, but it does not fully determine the spatial extent or contour shape of the reflected object. To provide dense geometric supervision, we introduce boundary projection alignment, which compares the predicted reflection boundary with an expected boundary obtained by projecting the outside-mirror object through the mirror geometry.

Given the outside-mirror object mask $\mathbf{M}_{obj}$, we sample foreground pixels and lift them to 3D using the available depth and camera intrinsics. These points are reflected across the mirror plane using Eq.~\eqref{eq:reflect} and then reprojected into the image plane. The projected support inside the mirror is rasterized and softly dilated to obtain the expected reflection support. We then extract its soft contour as the projected target boundary $\mathbf{B}_{proj}\in[0,1]$.
\begin{figure}[t]
\centering
\includegraphics[width=0.92\linewidth]{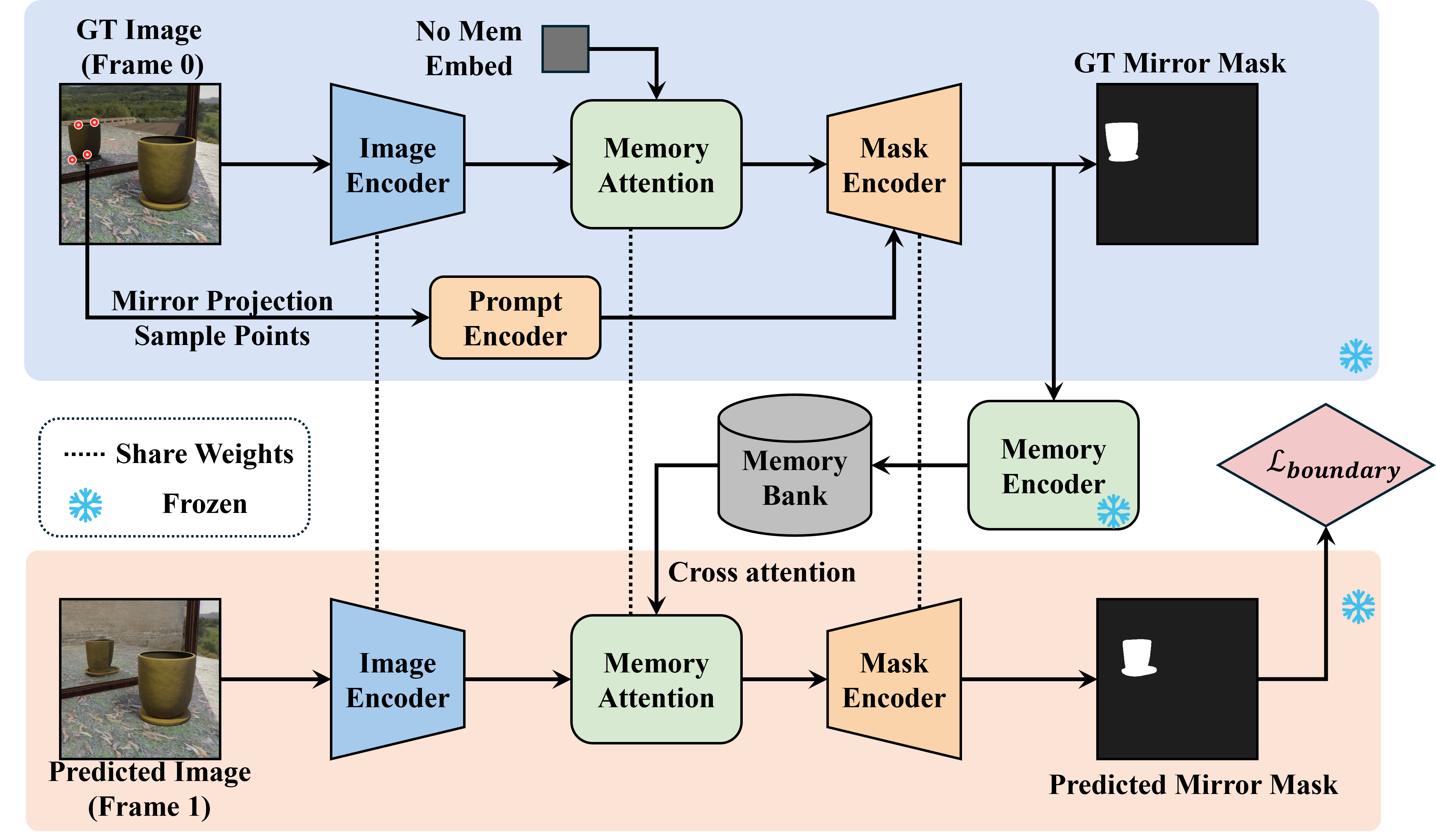}
\caption{\textbf{TwinTrack for source-boundary extraction.} The ground-truth image $\mathbf{X}_0$ and decoded prediction $\hat{\mathbf{X}}_0$ form a two-frame pseudo video. A frozen SAM2 propagates the ground-truth contour prompt to the prediction frame, producing a detached mask $\mathbf{M}_{sam2}$ that gates Sobel edges from $\hat{\mathbf{X}}_0$ and yields the differentiable source boundary $\hat{\mathbf{B}}$.}
\label{fig:twintrack}
\end{figure}
The predicted boundary is extracted from the decoded image prediction $\hat{\mathbf{X}}_0$. Directly segmenting the generated in-mirror object is unreliable during diffusion training, because the reflection may exhibit positional drift, shape distortion, or incomplete appearance. As shown in Fig.~\ref{fig:twintrack}, to obtain a stable supervision region,  we introduce \textit{TwinTrack}, which treats the ground-truth image $\mathbf{X}_0$ and the decoded prediction $\hat{\mathbf{X}}_0$ as a two-frame pseudo video. A frozen SAM2~\cite{sam2} propagates the ground-truth contour prompt from $\mathbf{X}_0$ to $\hat{\mathbf{X}}_0$, producing an in-mirror mask $\mathbf{M}_{sam2}$ that localizes the generated reflection.

Importantly, TwinTrack is used only as a detached spatial selector rather than as a shape target. We compute a differentiable edge map from the generated image:
\begin{equation}
    \hat{\mathbf{E}} =
    \sqrt{
    (S_x * \hat{\mathbf{X}}_{0,gray})^2
    +
    (S_y * \hat{\mathbf{X}}_{0,gray})^2
    },
\end{equation}
where $S_x$ and $S_y$ denote Sobel kernels. The TwinTrack mask is converted into a detached soft weight $\mathbf{W}_{sam}\in[0,1]$, and the predicted boundary is obtained as:
\begin{equation}
    \hat{\mathbf{B}} = \hat{\mathbf{E}} \odot \mathbf{W}_{sam}.
\end{equation}
Thus, gradients flow through the image-derived edge map $\hat{\mathbf{E}}$, while $\mathbf{W}_{sam}$ only determines where the boundary loss is evaluated.

We align the predicted boundary $\hat{\mathbf{B}}$ with the projected target boundary $\mathbf{B}_{proj}$ inside the mirror region. The loss combines a BCE term for local boundary activation and a Soft-Dice term for sparse contour overlap:
\begin{equation}
    \mathcal{L}_{boundary}
    =
    \lambda_{bce}\mathcal{L}_{bce}
    +
    \lambda_{dice}\mathcal{L}_{dice}.
    \label{eq:boundarydice}
\end{equation}
Combining sparse epipolar correspondence and dense boundary projection alignment, the final geometric loss is defined as:
\begin{equation}
    \mathcal{L}_{geo}
    =
    \lambda_{epi}\mathcal{L}_{epi}
    +
    \lambda_{boundary}\mathcal{L}_{boundary}.
\end{equation}

\subsection{Perceptual Loss}
\label{sec:perc}

Geometric supervision constrains where the reflection appears and how its contour aligns with the mirror-induced projection, but it does not guarantee that the reflected content remains perceptually faithful. A reflection may be geometrically well aligned while still suffering from identity drift, color shifts, texture changes, or inconsistent illumination. At the same time, exact pixel-level matching can be overly restrictive because reflected appearance may differ locally in shading, contrast, and texture while remaining perceptually consistent with the source. We therefore introduce a \textbf{Perceptual Loss} $\mathcal{L}_{perc}$ that combines semantic consistency and lighting consistency to regularize the identity and appearance of generated reflections. These terms serve as perceptual and low-frequency illumination priors rather than exact optical constraints.

For semantic consistency, we use frozen DINOv2~\cite{dinov2} patch features to preserve the identity and appearance of reflected content. Let $\Phi(\cdot)$ denote the DINOv2 feature extractor. We compute semantic consistency at two scales: the mirror-local term $\mathcal{L}_{local}$ compares normalized patch features within the mirror region and provides the main constraint for reflected-content fidelity, while the global term $\mathcal{L}_{global}$ compares full-image features and acts as a scene-level regularizer. The semantic consistency loss is:
\begin{equation}
    \mathcal{L}_{sem}
    =
    \lambda_{global}\mathcal{L}_{global}
    +
    \lambda_{local}\mathcal{L}_{local}.
    \label{eq:sem_total}
\end{equation}

For lighting consistency, we regularize geometry-aware photometric coherence in the reflected region. A frozen dense prediction model, Lotus~\cite{lotus}, estimates depth and surface normals from the decoded prediction $\hat{\mathbf{X}}_0$. The depth-normal term $\mathcal{L}_{dn}$ encourages geometry-compatible local structure, while the spherical-harmonics term $\mathcal{L}_{sh}$ captures low-frequency illumination consistency under the estimated normals. Specifically, we fit second-order spherical harmonics in the mirror region and use the resulting normal-conditioned shading representation to define $\mathcal{L}_{sh}$, without requiring exact pixel-level lighting reproduction. Here, spherical harmonics are used as a compact descriptor of low-frequency scene illumination rather than as a reflectance model of the mirror surface. The lighting consistency loss is:
\begin{equation}
    \mathcal{L}_{light}
    =
    \lambda_{dn}\mathcal{L}_{dn}
    +
    \lambda_{sh}\mathcal{L}_{sh}.
    \label{eq:light_total}
\end{equation}

The final perceptual loss combines semantic and lighting consistency:
\begin{equation}
    \mathcal{L}_{perc}
    =
    \lambda_{sem}\mathcal{L}_{sem}
    +
    \lambda_{light}\mathcal{L}_{light}.
    \label{eq:perc_total}
\end{equation}
\subsection{Training Strategy}
\label{sec:training}

PhysReflect augments the standard diffusion denoising objective with training-time physical supervision. The overall training loss is:
\begin{equation}
\mathcal{L}_{total}
=
\mathcal{L}_{MSE}
+
w(t)
\left(
\lambda_{geo}\mathcal{L}_{geo}
+
\lambda_{perc}\mathcal{L}_{perc}
\right),
\label{eq:total_loss}
\end{equation}
where $\mathcal{L}_{MSE}$ denotes the standard noise-prediction loss, and $w(t)$ is a timestep-dependent weight for pixel-space physical supervision. We follow the three-stage training protocol of MirrorVerse~\cite{mirrorverse} and apply these losses on top of this base training procedure.

Both geometric and perceptual losses are computed on the decoded prediction $\hat{\mathbf{X}}_0$ or structures derived from it. Their reliability therefore depends on the quality of the decoded image prediction at each diffusion timestep. At high-noise timesteps, $\hat{\mathbf{X}}_0$ may contain unstable structures and inaccurate photometric cues, making geometric matching, boundary extraction, semantic feature comparison, and lighting estimation unreliable. We therefore introduce an annealed supervision scheme, where $w(t)$ down-weights losses at unreliable high-noise timesteps and emphasizes low-noise predictions where pixel-space physical measurements are more stable.

All auxiliary estimators, matchers, and cached targets are used only during training. At inference time, PhysReflect keeps the original BrushNet inputs and sampling procedure unchanged, introducing no additional inference overhead. Details of $w(t)$, single-step decoding, and caching are provided in Appendix.
\section{Experiments}

\subsection{Experimental Setup}
\label{sec:setup}
\subsubsection{Datasets and Evaluation Metrics.}
We evaluate PhysReflect on the synthetic MirrorBenchV2 benchmark derived from SynMirrorV2~\cite{mirrorverse} and the real-world MSD dataset~\cite{msd}. SynMirrorV2 contains 201{,}610 Blender-rendered mirror-scene samples built from 66{,}062 target objects, and MirrorBenchV2 is the evaluation split derived from SynMirrorV2. We report results on its single-object and multi-object splits. MSD contains 4{,}018 real indoor mirror images with cluttered layouts, diverse mirror shapes, and natural illumination, and is used to test real-world generalization.

We evaluate mirror reflections using three categories of metrics:
\begin{itemize}
    \item \textbf{Visual Quality:} Measures pixel-level fidelity within the mirror region using PSNR~$\uparrow$, SSIM~$\uparrow$, and LPIPS~$\downarrow$~\cite{lpips}.
    \item \textbf{Semantic Consistency:} Assesses perceptual preservation of object identity and appearance via CLIP Similarity~$\uparrow$~\cite{clip} and DINO Similarity~$\uparrow$~\cite{dinov2}.
    \item \textbf{Physical Plausibility:} Evaluates geometry and illumination consistency. IoU~$\uparrow$ and Boundary F1~$\uparrow$ measure mask and contour alignment, m\_IoU~$\uparrow$ and m\_BF1~$\uparrow$ average multi-object results, and SH\_Err~$\downarrow$ measures low-frequency illumination consistency using second-order spherical harmonics.
\end{itemize}
\subsubsection{Implementation Details.}
For a fair comparison, we follow the method settings of prior work and fine-tune the pretrained Stable Diffusion v1.5~\cite{rombach2022stablediffusion} model with the Adam optimizer, using a learning rate of $1 \times 10^{-5}$ and a batch size of 4. Training is conducted in FP32 precision on four NVIDIA H100 GPUs, with 60{,}000, 30{,}000, and 20{,}000 steps for the single-object, multi-object, and real-world data stages, respectively. All methods are evaluated at $512 \times 512$ resolution using the same sampling strategy, prompt inputs, four random seeds per sample, and the best unmasked-region SSIM selection protocol of MirrorVerse~\cite{mirrorverse}. All ablation experiments are run on two NVIDIA H100 GPUs. Additional experimental Setup details are provided in Appendix.

\subsection{Main Results on Synthetic Benchmarks}
\label{sec:quantitative}

We first evaluate PhysReflect on the synthetic MirrorBenchV2 benchmark, where paired ground truth reflections and geometric annotations are available. This setting allows us to jointly measure image fidelity, semantic preservation, and physical plausibility under controlled reflection geometry. We report results separately on the single-object split (Table~\ref{tab:single}) and the more challenging multi-object split (Table~\ref{tab:multi}) to expose how each method behaves as geometric complexity increases.

\begin{table*}[t]
\caption{\textbf{Single-object results on MirrorBenchV2.} Best results are in \textbf{bold}, second best in \underline{underlined}.}
\label{tab:single}
\centering
\scriptsize
\renewcommand{\arraystretch}{0.72}
\resizebox{\textwidth}{!}{
\begin{tabular}{l|ccc|cc|ccc}
\toprule
\multirow{2}{*}{Method} 
& \multicolumn{3}{c|}{Reflection Gen. Quality} 
& \multicolumn{2}{c|}{Semantic Consist.} 
& \multicolumn{3}{c}{Physical Plausibility} \\
& PSNR $\uparrow$ 
& SSIM $\uparrow$ 
& LPIPS $\downarrow$ 
& CLIP Sim $\uparrow$ 
& DINO Sim $\uparrow$ 
& IoU $\uparrow$ 
& Bound. F1 $\uparrow$ 
& SH\_Err $\downarrow$ \\
\midrule
BrushNet-FT~\cite{brushnet} 
& 18.44 & 0.76 & 0.121 & 25.75 & \underline{0.64} & 0.474 & 0.207 & 107.42 \\
MirrorFusion~\cite{mirrorfusion} 
& 18.38 & 0.76 & 0.122 & \textbf{25.89} & 0.63 & 0.409 & 0.182 & 51.48 \\
MirrorVerse~\cite{mirrorverse} 
& \underline{18.74} & \underline{0.77} & \underline{0.108} & 25.69 & \underline{0.64} & \underline{0.598} & \underline{0.316} & \underline{49.93} \\
\textbf{PhysReflect (Ours)} 
& \textbf{19.14} & \textbf{0.78} & \textbf{0.103} & \underline{25.87} & \textbf{0.68} & \textbf{0.671} & \textbf{0.406} & \textbf{15.33} \\
\bottomrule
\end{tabular}
}
\end{table*}

\begin{table*}[t]
\caption{\textbf{Multi-object results on MirrorBenchV2.} This split includes larger depth variation, occlusion, and complex layouts. Best results are in \textbf{bold}, second best in \underline{underlined}.}
\label{tab:multi}
\centering
\scriptsize
\setlength{\tabcolsep}{3pt}
\renewcommand{\arraystretch}{0.8}
\resizebox{\textwidth}{!}{
\begin{tabular}{l|ccc|cc|ccccc}
\toprule
\multirow{2}{*}{Method} & \multicolumn{3}{c|}{Reflection Gen. Quality} & \multicolumn{2}{c|}{Semantic Consist.} & \multicolumn{5}{c}{Physical Plausibility} \\
 & PSNR $\uparrow$ & SSIM $\uparrow$ & LPIPS $\downarrow$ & CLIP Sim $\uparrow$ & DINO Sim $\uparrow$ & IoU $\uparrow$ & Bound. F1 $\uparrow$ & m\_IoU $\uparrow$ & m\_BF1 $\uparrow$ & SH\_Err $\downarrow$ \\
\midrule
BrushNet-FT~\cite{brushnet} & 17.37 & \underline{0.752} & 0.143 & \textbf{27.22} & \underline{0.63} & 0.534 & 0.223 & 0.442 & 0.198 & 8.19 \\
MirrorFusion~\cite{mirrorfusion} & 17.15 & 0.734 & 0.144 & 26.64 & 0.59 & 0.541 & 0.249 & 0.457 & 0.225 & 5.28 \\
MirrorVerse~\cite{mirrorverse} & \underline{17.90} & 0.745 & \underline{0.119} & 26.04 & 0.61 & \underline{0.638} & \underline{0.324} & \underline{0.534} & \underline{0.289} & \underline{5.04} \\
\textbf{PhysReflect (Ours)} & \textbf{18.44} & \textbf{0.764} & \textbf{0.113} & \underline{26.76} & \textbf{0.64} & \textbf{0.734} & \textbf{0.467} & \textbf{0.621} & \textbf{0.417} & \textbf{4.51} \\
\bottomrule
\end{tabular}
}
\setlength{\tabcolsep}{6pt}
\end{table*}

\subsubsection{Single-Object Scenes.}
Table~\ref{tab:single} shows that physical supervision improves reflection correctness without sacrificing image fidelity. The gains over MirrorVerse are modest on reconstruction metrics but much larger on IoU, Boundary F1, and SH\_Err, which are the metrics most directly tied to our geometry and lighting losses. Together with the improvements in reconstruction and perceptual metrics and the human evaluation in Section~\ref{sec:user_study}, these results provide complementary evidence of improved reflection quality.

\begin{figure*}[!t]
\centering
\includegraphics[width=\textwidth,height=0.4\textheight,keepaspectratio]{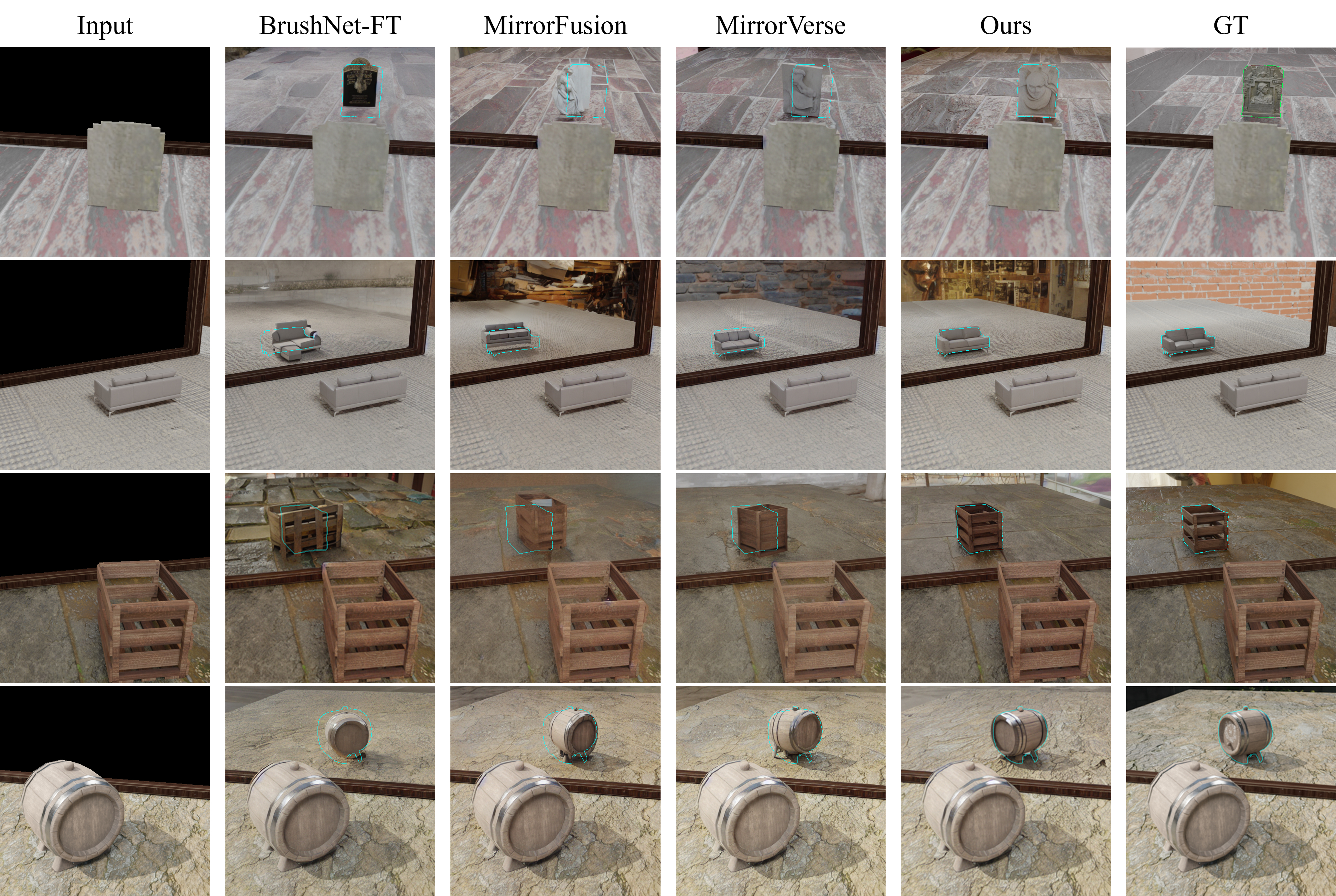}
\caption{\textbf{Single-object qualitative comparison on MirrorBenchV2.} PhysReflect improves projected-target alignment, identity preservation, and boundary quality. The final column is rendered GT.}
\label{fig:single_qualitative}
\end{figure*}

\subsubsection{Multi-Object Scenes.}
The multi-object split in Table~\ref{tab:multi} stresses the method with occlusion, depth variation, and multiple reflected instances. The advantage of PhysReflect becomes stronger in this setting, especially on per-object geometry metrics. This supports the role of explicit physical constraints: when several objects compete for plausible mirror content, a data-only objective can produce locally convincing textures but loses instance-level placement, whereas our losses keep each reflection tied to its corresponding projected geometry. The high CLIP score of BrushNet-FT further illustrates that text-image alignment alone is not a reliable proxy for physically correct reflection layout.

\begin{figure*}[!t]
\centering
\includegraphics[width=\textwidth,height=0.4\textheight,keepaspectratio]{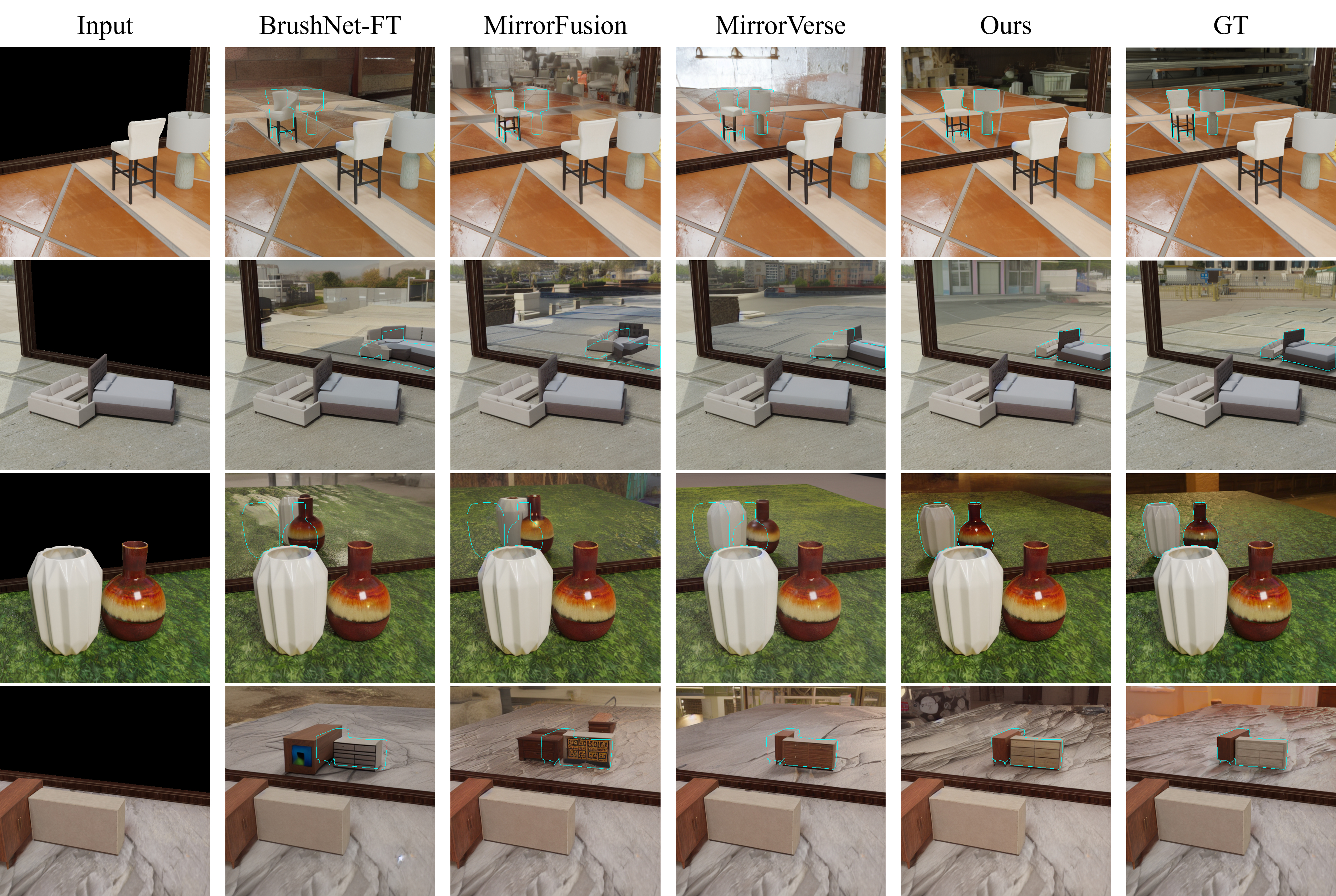}
\caption{\textbf{Multi-object qualitative comparison on MirrorBenchV2.} PhysReflect reduces missing, duplicated, or misplaced reflected objects under occlusion and depth variation. The final column is rendered GT.}
\label{fig:multi_qualitative}
\end{figure*}

\subsection{Generalization to Real-World Scenes}
\label{sec:real}

We next evaluate on MSD to test whether the learned physical constraints transfer to real mirror scenes with natural illumination, clutter, diverse mirror shapes, and capture noise. Since MSD is a real-world dataset, depth maps and camera intrinsics are estimated rather than rendered, and metrics are computed against the captured target image within the mirror region. We therefore treat this image as a reference target, not as deterministic rendered GT. We compare direct evaluation and MSD-adapted settings for both PhysReflect and MirrorVerse.

\begin{table}[t]
\caption{\textbf{Real-world generalization on MSD.} ``FT'' denotes MSD fine-tuning; methods without ``FT'' are evaluated directly on MSD. Best results are in \textbf{bold}, second best in \underline{underlined}.}
\label{tab:msd}
\centering
\footnotesize
\setlength{\tabcolsep}{2pt}
\renewcommand{\arraystretch}{0.90}
\begin{tabular}{lcccccc}
\toprule
Method & PSNR $\uparrow$ & SSIM $\uparrow$ & LPIPS $\downarrow$ & CLIP $\uparrow$ & DINO $\uparrow$ & SH\_Err $\downarrow$ \\
\midrule
MirrorVerse & 16.76 & 0.843 & 0.132 & \underline{26.90} & 0.48 & 128.74 \\
MirrorVerse-FT & 16.61 & 0.850 & \underline{0.119} & \textbf{27.67} & 0.51 & 96.47 \\
\textbf{PhysReflect} & \underline{16.84} & \underline{0.851} & 0.129 & 26.41 & \underline{0.52} & \underline{75.95} \\
\textbf{PhysReflect-FT} & \textbf{17.11} & \textbf{0.860} & \textbf{0.114} & 26.72 & \textbf{0.58} & \textbf{65.69} \\
\bottomrule
\end{tabular}
\setlength{\tabcolsep}{6pt}
\renewcommand{\arraystretch}{1.0}
\end{table}

Table~\ref{tab:msd} shows the same trend on real scenes. PhysReflect does not maximize CLIP similarity, but it improves image fidelity, DINO-based visual consistency, and SH\_Err, indicating better preservation of object appearance and lighting structure. The non-finetuned PhysReflect row already approaches or exceeds the MSD-finetuned MirrorVerse baseline on several metrics, which suggests that the proposed supervision transfers beyond the synthetic distribution rather than only memorizing dataset-specific reflection patterns.

\begin{figure}[!htbp]
\centering
\includegraphics[width=\columnwidth]{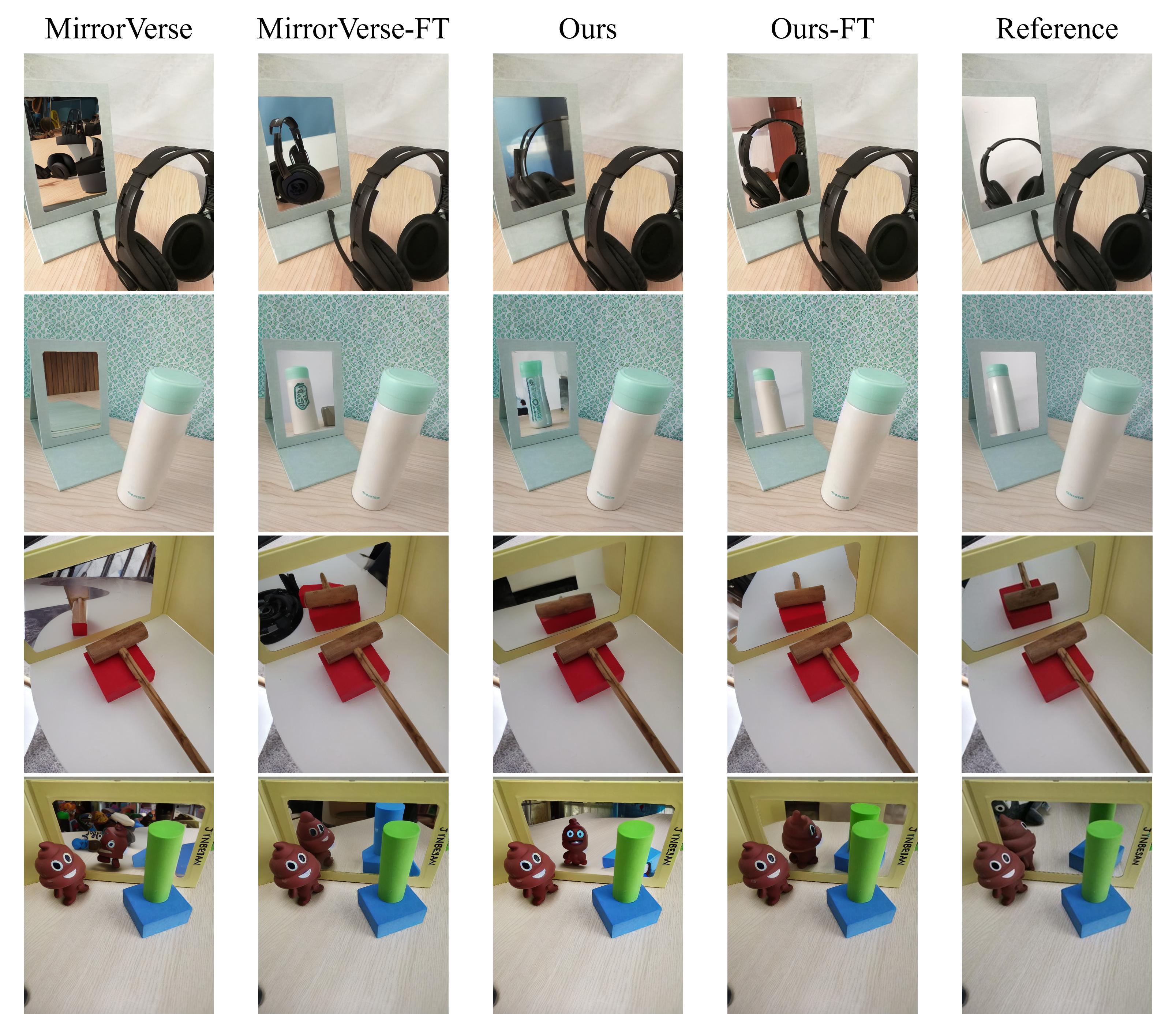}
\caption{\textbf{Qualitative comparison on MSD.} Reference denotes the captured target image used for visual comparison; MSD depth and camera intrinsics are estimated, so this target is not deterministic rendered GT.}
\label{fig:msd_qualitative}
\end{figure}

\subsection{Qualitative Results}
\label{sec:qualitative}

The synthetic comparisons in Figures~\ref{fig:single_qualitative} and~\ref{fig:multi_qualitative} show that PhysReflect reduces spatial drift, missing or duplicated reflected objects, and boundary mismatch in both single-object and multi-object scenes. The MSD visualization in Figure~\ref{fig:msd_qualitative} further illustrates the same behavior on real images, where the final column is denoted as Reference because real captured reflections are used for visual comparison rather than deterministic rendered ground truth.

\FloatBarrier

\subsection{User Study}
\label{sec:user_study}
\begin{figure}[t]
\centering
\includegraphics[width=0.90\columnwidth]{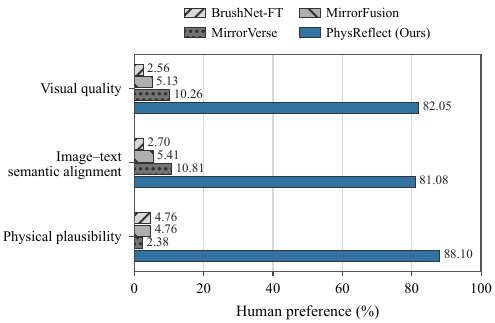}
\caption{\textbf{User study.} Distribution of recorded preferences (\%) among consensus samples in a four-way comparison by nine evaluators. Higher is better.}
\label{fig:user_study}
\end{figure}

To complement the automated metrics, we conducted a blind human perceptual study with nine evaluators. For each evaluated sample and criterion, participants performed a four-way forced-choice comparison among PhysReflect, MirrorVerse~\cite{mirrorverse}, MirrorFusion~\cite{mirrorfusion}, and BrushNet-FT~\cite{brushnet}. A strict-majority preference was recorded when one method received at least five of the nine votes. Strict-majority consensus was reached for 58.5\%, 55.5\%, and 84.0\% of the 200 samples for visual quality, image--text semantic alignment, and perceived physical plausibility, respectively, with corresponding no-consensus rates of 41.5\%, 44.5\%, and 16.0\%. Figure~\ref{fig:user_study} reports the preference distribution among consensus samples, while sample-level and individual-vote statistics are provided in the supplementary material. Among consensus samples, PhysReflect receives the largest preference share across all three criteria, supporting that its improvements are perceptually observable rather than limited to automated metrics.

\subsection{Ablation Study}
\label{sec:ablation}

We distinguish between \textit{components}, which control where and when supervision is applied, and \textit{supervisions}, which define what physical signal is enforced. The components are TwinTrack and \textit{Annealed Supervision}; the supervisions are epipolar supervision $\mathcal{L}_{epi}$, boundary supervision $\mathcal{L}_{boundary}$, semantic supervision $\mathcal{L}_{sem}$, and lighting supervision $\mathcal{L}_{light}$. Alternative Annealed Supervision variants are reported in the appendix.

\begin{table*}[t]
\caption{\textbf{Ablation of components and supervisions on the single-object split of MirrorBenchV2.} Checkmarks indicate enabled components or losses. Annealed Supervision denotes the final inverse-timestep reweighting strategy. Best results are in \textbf{bold}.}
\label{tab:ablation}
\centering
\scriptsize
{\renewcommand{\arraystretch}{0.5}
\resizebox{\textwidth}{!}{
\begin{tabular}{cc|cccc|ccc|cc|ccc}
\toprule
\multicolumn{6}{c|}{Method} & \multicolumn{8}{c}{Metric} \\
\cmidrule(lr){1-6}\cmidrule(lr){7-14}
\multicolumn{2}{c|}{Component} & \multicolumn{4}{c|}{Supervision} & \multicolumn{3}{c|}{Reflection Gen. Quality} & \multicolumn{2}{c|}{Semantic Consist.} & \multicolumn{3}{c}{Physical Plausibility} \\
\cmidrule(lr){1-14}
\shortstack{Annealed Supervision} & TwinTrack & $\mathcal{L}_{epi}$ & $\mathcal{L}_{boundary}$ & $\mathcal{L}_{sem}$ & $\mathcal{L}_{light}$ & PSNR $\uparrow$ & SSIM $\uparrow$ & LPIPS $\downarrow$ & CLIP Sim $\uparrow$ & DINO Sim $\uparrow$ & IoU $\uparrow$ & Bound. F1 $\uparrow$ & SH\_Err $\downarrow$ \\
\midrule
 &  &  &  &  &  & 18.33 & 0.76 & 0.109 & 25.68 & 0.61 & 0.554 & 0.296 & 58.41 \\
$\checkmark$ &  & $\checkmark$ &  &  &  & 18.24 & 0.77 & 0.112 & 25.47 & 0.65 & 0.581 & 0.347 & 59.36 \\
$\checkmark$ &  &  & $\checkmark$ &  &  & 18.44 & 0.77 & 0.110 & 25.82 & 0.62 & 0.541 & 0.237 & 49.24 \\
$\checkmark$ & $\checkmark$ &  & $\checkmark$ &  &  & 18.41 & 0.78 & 0.108 & 25.83 & 0.66 & 0.607 & 0.351 & 58.03 \\
$\checkmark$ & $\checkmark$ & $\checkmark$ & $\checkmark$ &  &  & 18.27 & \textbf{0.81} & 0.111 & 25.47 & 0.67 & 0.611 & \textbf{0.357} & 56.33 \\
 & $\checkmark$ & $\checkmark$ & $\checkmark$ &  &  & 15.28 & 0.69 & 0.153 & 24.66 & 0.51 & 0.352 & 0.189 & 114.42 \\
$\checkmark$ &  &  &  & $\checkmark$ &  & \textbf{19.04} & 0.78 & 0.108 & \textbf{25.91} & 0.68 & 0.593 & 0.318 & 20.51 \\
$\checkmark$ &  &  &  &  & $\checkmark$ & 18.85 & 0.78 & 0.109 & 25.76 & 0.66 & 0.599 & 0.316 & \textbf{16.84} \\
$\checkmark$ &  &  &  & $\checkmark$ & $\checkmark$ & 18.97 & 0.78 & 0.108 & 25.88 & 0.68 & 0.604 & 0.322 & 16.97 \\
 &  &  &  & $\checkmark$ & $\checkmark$ & 18.21 & 0.74 & 0.114 & 25.85 & \textbf{0.69} & 0.515 & 0.289 & 87.24 \\
$\checkmark$ & $\checkmark$ & $\checkmark$ & $\checkmark$ & $\checkmark$ & $\checkmark$ & 18.93 & 0.78 & \textbf{0.107} & 25.89 & 0.67 & \textbf{0.621} & 0.351 & 17.25 \\
\bottomrule
\end{tabular}
}
}
\end{table*}

Table~\ref{tab:ablation} separates the effect of supervision scheduling from that of supervision content. Annealed Supervision improves the balance between feature consistency, reconstruction, and geometry, indicating that pixel-space supervision is most effective at reliable denoising stages. TwinTrack further strengthens boundary supervision by providing a cleaner detached selector for the generated reflection contour.

The supervision terms are complementary: epipolar and boundary losses improve geometric plausibility, semantic supervision preserves reflected content, and lighting supervision regularizes shading coherence. No single term dominates every metric, whereas the full model provides the strongest overall balance between physical plausibility and visual quality. 

\begin{figure}[!htbp]
\centering
\includegraphics[width=\columnwidth,height=0.3\textheight,keepaspectratio]{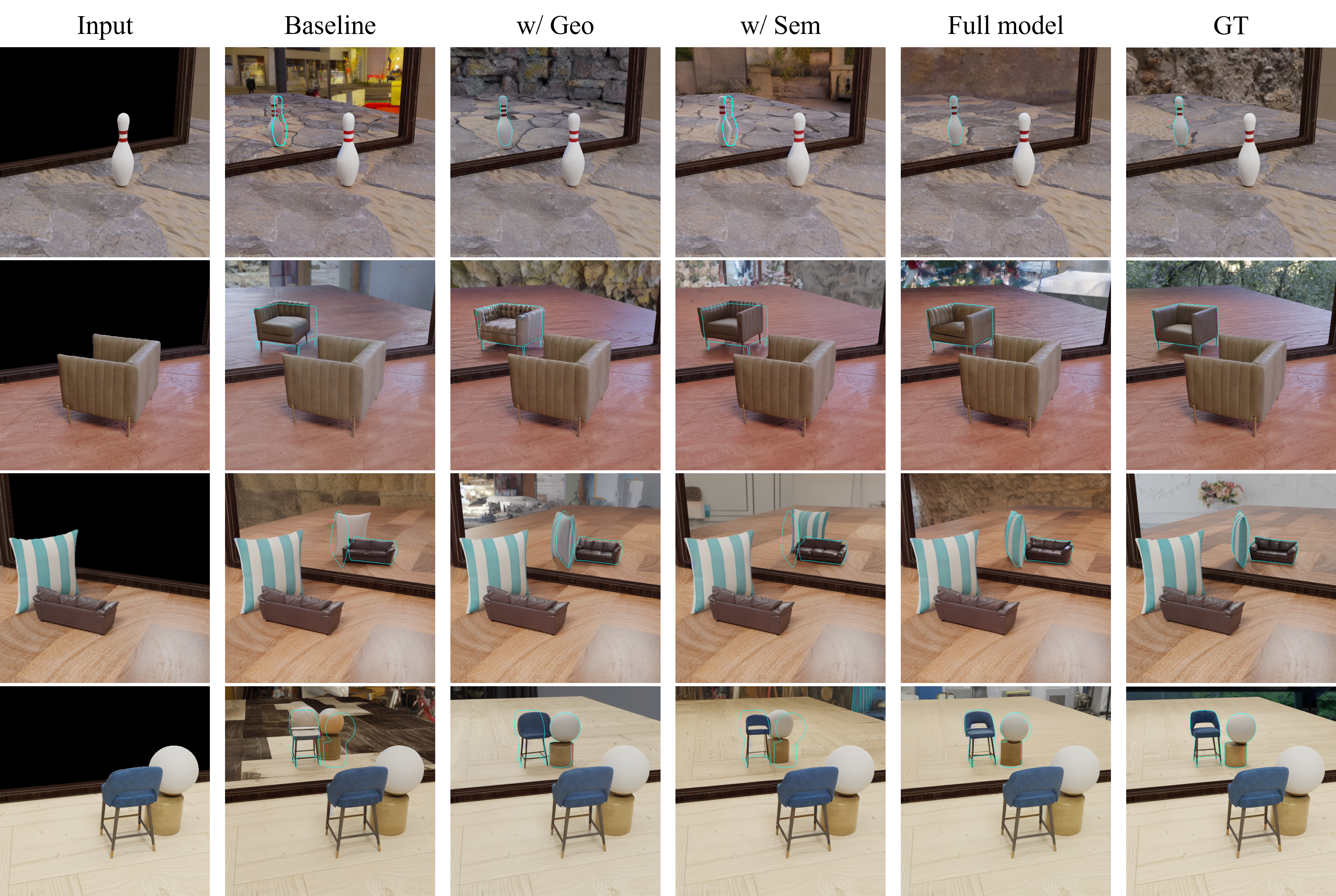}
\caption{\textbf{Geometry and semantic ablation.} Columns are Input, Baseline, w/ Geo, w/ Sem, Full, and GT, where w/ Geo and w/ Sem denote adding $\mathcal{L}_{epi}+\mathcal{L}_{boundary}$ and $\mathcal{L}_{sem}$ to the Baseline, respectively. Full denotes the complete model with all proposed components and losses.}
\label{fig:geo_sem_ablation}
\end{figure}

\begin{figure}[!htbp]
\centering
\includegraphics[width=\columnwidth,height=0.2\textheight,keepaspectratio]{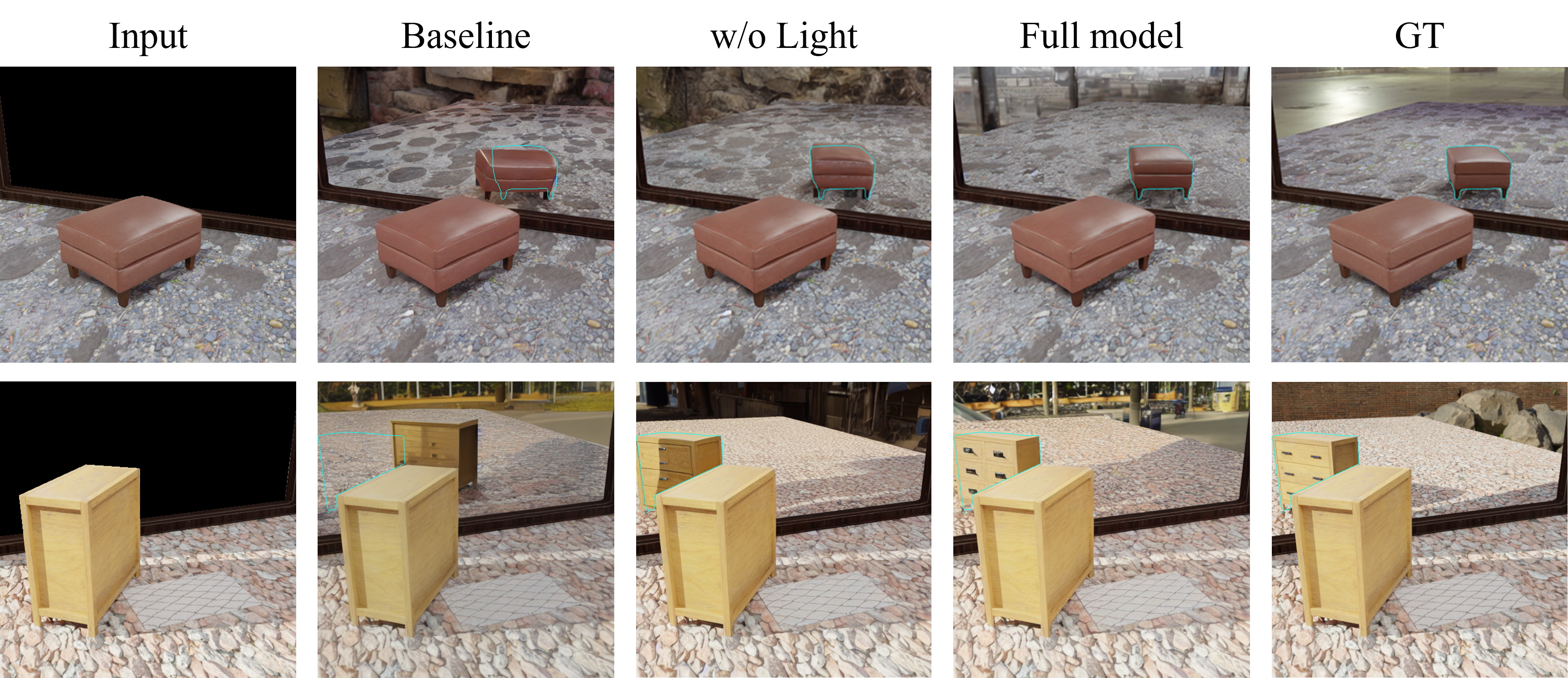}
\caption{\textbf{Lighting ablation.} The w/o lighting result denotes the Full model without the lighting loss, while the Full model uses all proposed losses. The lighting loss improves highlights, shading transitions, and mirror-region illumination coherence.}
\label{fig:light_ablation}
\end{figure}

The ablation visualizations in Figures~\ref{fig:geo_sem_ablation} and~\ref{fig:light_ablation} provide complementary evidence for the design choices. The geometry and semantic comparison shows that geometry supervision mainly anchors reflected content to the projected support, while semantic supervision improves object identity and texture preservation. As shown in the lighting ablation, the lighting loss improves highlights and the transition of reflection shading without changing the inference-time input.

\subsection{Efficiency}
\label{sec:efficiency}

\begin{table}[t]
\caption{\textbf{Inference cost comparison.} Average time over 100 images and peak memory for one $512 \times 512$ image on a single H100 GPU using FP32 precision. Lower is better.}
\label{tab:efficiency}
\centering
\footnotesize
\setlength{\tabcolsep}{1pt}
\renewcommand{\arraystretch}{0.85}
\begin{tabular}{l|ccc|ccc}
\toprule
\multirow{2}{*}{Method} & \multicolumn{3}{c|}{DDIM 50} & \multicolumn{3}{c}{UniPC 15} \\
 & S-Time $\downarrow$ & M-Time $\downarrow$ & Mem. $\downarrow$ & S-Time $\downarrow$ & M-Time $\downarrow$ & Mem. $\downarrow$ \\
\midrule
BrushNet-FT & 7.12 & 7.06 & 9.76 & 1.28 & 1.25 & 9.97 \\
MirrorVerse & 7.31 & 7.29 & 9.89 & 1.45 & 1.42 & 10.21 \\
\textbf{PhysReflect} & 7.31 & 7.23 & 9.82 & 1.45 & 1.38 & 10.03 \\
\bottomrule
\end{tabular}
\setlength{\tabcolsep}{6pt}
\renewcommand{\arraystretch}{1.0}
\end{table}
PhysReflect is designed to shift additional cost to training rather than inference. Table~\ref{tab:efficiency} confirms this behavior: because TwinTrack, DINOv2, and Lotus are frozen training-time modules, the deployed model keeps the same inputs and nearly the same computation graph as the BrushNet-based generator. The measured latency and memory remain essentially unchanged under both DDIM and UniPC sampling, so the physical gains above do not come from extra inference-time computation.

\section{Conclusion}
We presented PhysReflect, a geometry- and perception-guided diffusion framework that applies mirror-specific physical supervision to decoded clean-image predictions in pixel space during training. Through complementary geometric and perceptual objectives with reliability-aware annealed supervision, PhysReflect improves physical plausibility on synthetic and real-world benchmarks while leaving the inference procedure unchanged and preserving perceptual quality and efficiency.

\paragraph{Limitations.}
Our method is currently designed for predominantly planar-mirror scenes and relies on estimated geometric priors, whose errors may weaken supervision in challenging real-world cases. The synthetic-to-real domain gap also remains a limitation. The frozen training-time priors introduce additional training cost but do not affect inference efficiency. Representative failure cases are provided in the Appendix.

\FloatBarrier

\bibliographystyle{ACM-Reference-Format}
\bibliography{references}

@article{ho2020ddpm,
  title={Denoising diffusion probabilistic models},
  author={Ho, Jonathan and Jain, Ajay and Abbeel, Pieter},
  journal={Advances in neural information processing systems},
  volume={33},
  pages={6840--6851},
  year={2020}
}

@inproceedings{rombach2022stablediffusion,
  title={High-Resolution Image Synthesis with Latent Diffusion Models},
  author={Rombach, Robin and Blattmann, Andreas and Lorenz, Dominik and Esser, Patrick and Ommer, Bjorn},
  booktitle={Proceedings of the IEEE/CVF Conference on Computer Vision and Pattern Recognition},
  pages={10684--10695},
  year={2022}
}

@inproceedings{podell2023sdxl,
  title={Sdxl: Improving latent diffusion models for high-resolution image synthesis},
  author={Podell, Dustin and English, Zion and Lacey, Kyle and Blattmann, Andreas and Dockhorn, Tim and M{\"u}ller, Jonas and Penna, Joe and Rombach, Robin},
  booktitle={International Conference on Learning Representations},
  volume={2024},
  pages={1862--1874},
  year={2024}
}

@inproceedings{controlnet,
  title={Adding conditional control to text-to-image diffusion models},
  author={Zhang, Lvmin and Rao, Anyi and Agrawala, Maneesh},
  booktitle={Proceedings of the IEEE/CVF international conference on computer vision},
  pages={3836--3847},
  year={2023}
}

@inproceedings{mo2024freecontrol,
  title={Freecontrol: Training-free spatial control of any text-to-image diffusion model with any condition},
  author={Mo, Sicheng and Mu, Fangzhou and Lin, Kuan Heng and Liu, Yanli and Guan, Bochen and Li, Yin and Zhou, Bolei},
  booktitle={Proceedings of the IEEE/CVF conference on computer vision and pattern recognition},
  pages={7465--7475},
  year={2024}
}

@inproceedings{parihar2024precisecontrol,
  title={Balancing act: Distribution-guided debiasing in diffusion models},
  author={Parihar, Rishubh and Bhat, Abhijnya and Basu, Abhipsa and Mallick, Saswat and Kundu, Jogendra Nath and Babu, R Venkatesh},
  booktitle={Proceedings of the IEEE/CVF conference on computer vision and pattern recognition},
  pages={6668--6678},
  year={2024}
}

@inproceedings{parihar2025text2place,
  title={Text2place: Affordance-aware text guided human placement},
  author={Parihar, Rishubh and Gupta, Harsh and VS, Sachidanand and Babu, R Venkatesh},
  booktitle={European Conference on Computer Vision},
  pages={57--77},
  year={2024},
  organization={Springer}
}

@inproceedings{brushnet,
  title={Brushnet: A plug-and-play image inpainting model with decomposed dual-branch diffusion},
  author={Ju, Xuan and Liu, Xian and Wang, Xintao and Bian, Yuxuan and Shan, Ying and Xu, Qiang},
  booktitle={European Conference on Computer Vision},
  pages={150--168},
  year={2024},
  organization={Springer}
}

@inproceedings{sarkar2024shadows,
  title={Shadows don't lie and lines can't bend! generative models don't know projective geometry... for now},
  author={Sarkar, Ayush and Mai, Hanlin and Mahapatra, Amitabh and Lazebnik, Svetlana and Forsyth, David A and Bhattad, Anand},
  booktitle={Proceedings of the IEEE/CVF conference on computer vision and pattern recognition},
  pages={28140--28149},
  year={2024}
}

@article{upadhyay2023perspective,
  title={Enhancing diffusion models with 3d perspective geometry constraints},
  author={Upadhyay, Rishi and Zhang, Howard and Ba, Yunhao and Yang, Ethan and Gella, Blake and Jiang, Sicheng and Wong, Alex and Kadambi, Achuta},
  journal={ACM Transactions on Graphics (TOG)},
  volume={42},
  number={6},
  pages={1--15},
  year={2023},
  publisher={ACM New York, NY, USA}
}

@inproceedings{objectdrop,
  title={Objectdrop: Bootstrapping counterfactuals for photorealistic object removal and insertion},
  author={Winter, Daniel and Cohen, Matan and Fruchter, Shlomi and Pritch, Yael and Rav-Acha, Alex and Hoshen, Yedid},
  booktitle={European Conference on Computer Vision},
  pages={112--129},
  year={2024},
  organization={Springer}
}

@misc{sd35,
  title={Stable Diffusion 3.5},
  author={{Stability AI}},
  year={2024},
  howpublished={\url{https://stability.ai/news/introducing-stable-diffusion-3-5}}
}

@misc{flux,
  title={{FLUX}},
  author={{Black Forest Labs}},
  year={2024},
  howpublished={\url{https://blackforestlabs.ai/announcing-black-forest-labs/}}
}

@inproceedings{mirrorfusion,
  title={Reflecting reality: Enabling diffusion models to produce faithful mirror reflections},
  author={Dhiman, Ankit and Shah, Manan and Parihar, Rishubh and Bhalgat, Yash and Boregowda, Lokesh R and Babu, R Venkatesh},
  booktitle={2025 International Conference on 3D Vision (3DV)},
  pages={824--834},
  year={2025},
  organization={IEEE}
}

@inproceedings{mirrorverse,
  title={Mirrorverse: Pushing diffusion models to realistically reflect the world},
  author={Dhiman, Ankit and Shah, Manan and Babu, R Venkatesh},
  booktitle={Proceedings of the Computer Vision and Pattern Recognition Conference},
  pages={11239--11249},
  year={2025}
}

@inproceedings{msd,
  title={Where is my mirror?},
  author={Yang, Xin and Mei, Haiyang and Xu, Ke and Wei, Xiaopeng and Yin, Baocai and Lau, Rynson WH},
  booktitle={Proceedings of the IEEE/CVF international conference on computer vision},
  pages={8809--8818},
  year={2019}
}

@article{avrahami2023blended,
  title={Blended latent diffusion},
  author={Avrahami, Omri and Fried, Ohad and Lischinski, Dani},
  journal={ACM transactions on graphics (TOG)},
  volume={42},
  number={4},
  pages={1--11},
  year={2023},
  publisher={ACM New York, NY, USA}
}

@inproceedings{lugmayr2022repaint,
  title={Repaint: Inpainting using denoising diffusion probabilistic models},
  author={Lugmayr, Andreas and Danelljan, Martin and Romero, Andres and Yu, Fisher and Timofte, Radu and Van Gool, Luc},
  booktitle={Proceedings of the IEEE/CVF conference on computer vision and pattern recognition},
  pages={11461--11471},
  year={2022}
}

@inproceedings{lin2020pmd,
  title={Progressive mirror detection},
  author={Lin, Jiaying and Wang, Guodong and Lau, Rynson WH},
  booktitle={Proceedings of the IEEE/CVF conference on computer vision and pattern recognition},
  pages={3697--3705},
  year={2020}
}

@article{dinov2,
  title={Dinov2: Learning robust visual features without supervision},
  author={Oquab, Maxime and Darcet, Timoth{\'e}e and Moutakanni, Th{\'e}o and Vo, Huy and Szafraniec, Marc and Khalidov, Vasil and Fernandez, Pierre and Haziza, Daniel and Massa, Francisco and El-Nouby, Alaaeldin and others},
  journal={arXiv preprint arXiv:2304.07193},
  year={2023}
}

@inproceedings{clip,
  title={Learning transferable visual models from natural language supervision},
  author={Radford, Alec and Kim, Jong Wook and Hallacy, Chris and Ramesh, Aditya and Goh, Gabriel and Agarwal, Sandhini and Sastry, Girish and Askell, Amanda and Mishkin, Pamela and Clark, Jack and others},
  booktitle={International conference on machine learning},
  pages={8748--8763},
  year={2021},
  organization={PmLR}
}

@inproceedings{sam,
  title={Segment anything},
  author={Kirillov, Alexander and Mintun, Eric and Ravi, Nikhila and Mao, Hanzi and Rolland, Chloe and Gustafson, Laura and Xiao, Tete and Whitehead, Spencer and Berg, Alexander C and Lo, Wan-Yen and others},
  booktitle={Proceedings of the IEEE/CVF international conference on computer vision},
  pages={4015--4026},
  year={2023}
}

@inproceedings{sam2,
  title={Sam 2: Segment anything in images and videos},
  author={Ravi, Nikhila and Gabeur, Valentin and Hu, Yuan-Ting and Hu, Ronghang and Ryali, Chaitanya and Ma, Tengyu and Khedr, Haitham and R{\"a}dle, Roman and Rolland, Chloe and Gustafson, Laura and others},
  booktitle={International Conference on Learning Representations},
  volume={2025},
  pages={28085--28128},
  year={2025}
}

@inproceedings{lotus,
  title={Lotus: Diffusion-based visual foundation model for high-quality dense prediction},
  author={He, Jing and Li, Haodong and Yin, Wei and Liang, Yixun and Li, Leheng and Zhou, Kaiqiang and Zhang, Hongbo and Liu, Bingbing and Chen, YingCong},
  booktitle={International Conference on Learning Representations},
  volume={2025},
  pages={89454--89467},
  year={2025}
}

@article{unilumos,
  title={UniLumos: Fast and Unified Image and Video Relighting with Physics-Plausible Feedback},
  author={Liu, Pengwei and Yuan, Hangjie and Dong, Bo and Xing, Jiazheng and Wang, Jinwang and Zhao, Rui and Chen, Weihua and Wang, Fan},
  journal={Advances in Neural Information Processing Systems},
  volume={38},
  pages={82052--82080},
  year={2026}
}

@article{pilight,
  title={PI-Light: Physics-Inspired Diffusion for Full-Image Relighting},
  author={Liang, Zhexin and Chen, Zhaoxi and Chen, Yongwei and Wei, Tianyi and Wang, Tengfei and Pan, Xingang},
  journal={arXiv preprint arXiv:2601.22135},
  year={2026}
}

@article{lowe2004sift,
  title={Distinctive image features from scale-invariant keypoints},
  author={Lowe, David G},
  journal={International journal of computer vision},
  volume={60},
  number={2},
  pages={91--110},
  year={2004},
  publisher={Springer}
}

@inproceedings{rublee2011orb,
  title={ORB: An efficient alternative to SIFT or SURF},
  author={Rublee, Ethan and Rabaud, Vincent and Konolige, Kurt and Bradski, Gary},
  booktitle={2011 International conference on computer vision},
  pages={2564--2571},
  year={2011},
  organization={Ieee}
}

@inproceedings{mei2022glass,
  title={Glass segmentation using intensity and spectral polarization cues},
  author={Mei, Haiyang and Dong, Bo and Dong, Wen and Yang, Jiaxi and Baek, Seung-Hwan and Heide, Felix and Peers, Pieter and Wei, Xiaopeng and Yang, Xin},
  booktitle={Proceedings of the IEEE/CVF conference on computer vision and pattern recognition},
  pages={12622--12631},
  year={2022}
}

@inproceedings{msnerf,
  title={Multi-space neural radiance fields},
  author={Yin, Ze-Xin and Qiu, Jiaxiong and Cheng, Ming-Ming and Ren, Bo},
  booktitle={Proceedings of the IEEE/CVF Conference on Computer Vision and Pattern Recognition},
  pages={12407--12416},
  year={2023}
}

@article{mirror3dgs,
  title={Mirror-3dgs: Incorporating mirror reflections into 3d gaussian splatting},
  author={Meng, Jiarui and Li, Haijie and Wu, Yanmin and Gao, Qiankun and Yang, Shuzhou and Zhang, Jian and Ma, Siwei},
  journal={arXiv preprint arXiv:2404.01168},
  year={2024}
}

@inproceedings{mirrorgaussian,
  title={Mirrorgaussian: Reflecting 3d gaussians for reconstructing mirror reflections},
  author={Liu, Jiayue and Tang, Xiao and Cheng, Freeman and Yang, Roy and Li, Zhihao and Liu, Jianzhuang and Huang, Yi and Lin, Jiaqi and Liu, Shiyong and Wu, Xiaofei and others},
  booktitle={European Conference on Computer Vision},
  pages={377--393},
  year={2024},
  organization={Springer}
}

@inproceedings{lpips,
  title={The unreasonable effectiveness of deep features as a perceptual metric},
  author={Zhang, Richard and Isola, Phillip and Efros, Alexei A and Shechtman, Eli and Wang, Oliver},
  booktitle={Proceedings of the IEEE conference on computer vision and pattern recognition},
  pages={586--595},
  year={2018}
}

@inproceedings{xu2023imagereward,
  title     = {ImageReward: Learning and Evaluating Human Preferences for Text-to-Image Generation},
  author    = {Xu, Jiazheng and Liu, Xiao and Wu, Yuchen and Tong, Yuxuan and Li, Qinkai and Ding, Ming and Tang, Jie and Dong, Yuxiao},
  booktitle = {Advances in Neural Information Processing Systems},
  volume    = {36},
  pages     = {15903--15935},
  year      = {2023}
}

@inproceedings{clark2024draft,
  title     = {Directly Fine-Tuning Diffusion Models on Differentiable Rewards},
  author    = {Clark, Kevin and Vicol, Paul and Swersky, Kevin and Fleet, David J.},
  booktitle = {International Conference on Learning Representations},
  year      = {2024}
}

@article{prabhudesai2023alignprop,
  title   = {Aligning Text-to-Image Diffusion Models with Reward Backpropagation},
  author  = {Prabhudesai, Mihir and Goyal, Anirudh and Pathak, Deepak and Fragkiadaki, Katerina},
  year    = {2023}
}

@inproceedings{yu2025repa,
  title     = {Representation Alignment for Generation: Training Diffusion Transformers Is Easier Than You Think},
  author    = {Yu, Sihyun and Kwak, Sangkyung and Jang, Huiwon and Jeong, Jongheon and Huang, Jonathan and Shin, Jinwoo and Xie, Saining},
  booktitle = {International Conference on Learning Representations},
  year      = {2025}
}

@article{ma2026pixelgen,
  title   = {PixelGen: Improving Pixel Diffusion with Perceptual Supervision},
  author  = {Ma, Zehong and Xu, Ruihan and Zhang, Shiliang},
  year    = {2026}
}

@inproceedings{huang2024epidiff,
  title     = {EpiDiff: Enhancing Multi-View Synthesis via Localized Epipolar-Constrained Diffusion},
  author    = {Huang, Zehuan and Wen, Hao and Dong, Junting and Wang, Yaohui and Li, Yangguang and Chen, Xinyuan and Cao, Yan-Pei and Liang, Ding and Qiao, Yu and Dai, Bo and Sheng, Lu},
  booktitle = {Proceedings of the IEEE/CVF Conference on Computer Vision and Pattern Recognition},
  pages     = {9784--9794},
  year      = {2024}
}

@inproceedings{liu2023zero123,
  title     = {Zero-1-to-3: Zero-shot One Image to 3D Object},
  author    = {Liu, Ruoshi and Wu, Rundi and Van Hoorick, Basile and Tokmakov, Pavel and Zakharov, Sergey and Vondrick, Carl},
  booktitle = {Proceedings of the IEEE/CVF International Conference on Computer Vision},
  pages     = {9298--9309},
  year      = {2023}
}

\end{document}